%% file: conference_101719.tex
\documentclass[conference]{IEEEtran}
\IEEEoverridecommandlockouts
\usepackage{cite}
\usepackage{amsmath,amssymb,amsfonts}
\usepackage{algorithmic}
\usepackage{graphicx}
\usepackage{textcomp}
\usepackage{xcolor}

\usepackage{hyperref}       %
\usepackage{colortbl}
\usepackage{multirow}
\usepackage{makecell}
\usepackage{color}

\usepackage{pifont}

\input{commands}

\def\BibTeX{{\rm B\kern-.05em{\sc i\kern-.025em b}\kern-.08em
    T\kern-.1667em\lower.7ex\hbox{E}\kern-.125emX}}
\begin{document}

\title{RealBench: Benchmarking Verilog Generation Models with Real-World IP Designs}

\author{\IEEEauthorblockN{
Pengwei Jin\IEEEauthorrefmark{\dag}\IEEEauthorrefmark{1}\thanks{\textsuperscript{\footnotesize \dag}Email: jinpengwei20z@ict.ac.cn}, 
Di Huang\IEEEauthorrefmark{1}, 
Chongxiao Li\IEEEauthorrefmark{1,2,5}, 
Shuyao Cheng\IEEEauthorrefmark{1}, 
Yang Zhao\IEEEauthorrefmark{1,2,5},
Xinyao Zheng\IEEEauthorrefmark{1,2,5}, \\
Jiaguo Zhu\IEEEauthorrefmark{1,3,5}, 
Shuyi Xing\IEEEauthorrefmark{1,3,5}, 
Bohan Dou\IEEEauthorrefmark{1,3,5}, 
Rui Zhang\IEEEauthorrefmark{1}, 
Zidong Du\IEEEauthorrefmark{1,4}, 
Qi Guo\IEEEauthorrefmark{1}, 
Xing Hu\IEEEauthorrefmark{*}\IEEEauthorrefmark{1,4}\thanks{\textsuperscript{\footnotesize *}Corresponding author. Email: huxing@ict.ac.cn}
}

\IEEEauthorblockA{\IEEEauthorrefmark{1}\textit{State Key Lab of Processors, Institute of Computing Technology, CAS}, Beijing, China}
\IEEEauthorblockA{\IEEEauthorrefmark{2}\textit{University of Chinese Academy of Sciences}, Beijing, China}
\IEEEauthorblockA{\IEEEauthorrefmark{3}\textit{University of Science and Technology of China}, Hefei, China}
\IEEEauthorblockA{\IEEEauthorrefmark{4}\textit{Shanghai Innovation Center for Processor Technologies}, Shanghai, China}
\IEEEauthorblockA{\IEEEauthorrefmark{5}\textit{Cambricon Technologies}}

\IEEEauthorblockA{\\\url{https://github.com/IPRC-DIP/RealBench}}
}

\maketitle

\begin{abstract}
The automatic generation of Verilog code using Large Language Models (LLMs) has garnered significant interest in hardware design automation. However, existing benchmarks for evaluating LLMs in Verilog generation fall short in replicating real-world design workflows due to their designs' simplicity, inadequate design specifications, and less rigorous verification environments. To address these limitations, we present \textit{\xname}, the first benchmark aiming at real-world IP-level Verilog generation tasks. \xname features complex, structured, real-world open-source IP designs, multi-modal and formatted design specifications, and rigorous verification environments, including $100\%$ line coverage testbenches and a formal checker. It supports both module-level and system-level tasks, enabling comprehensive assessments of LLM capabilities. 
Evaluations on various LLMs and agents reveal that even one of the best-performing LLMs, o1-preview, achieves only a $13.3\%$ pass@1 on module-level tasks and $0\%$ on system-level tasks,
highlighting the need for stronger Verilog generation models in the future.

\end{abstract}


\input{sections/1_introduction}

\input{sections/2_methods}

\input{sections/3_experiments}

\input{sections/4_relatedwork}

\input{sections/5_conclusion}

\bibliography{reference}

\end{document}

%% file: commands.tex
\newcommand{\ie}{\textit{i}.\textit{e}.\xspace}
\newcommand{\eg}{\textit{e}.\textit{g}.,\xspace}

\usepackage{xspace}
\newcommand{\xname}{RealBench\xspace}

\renewcommand{\IEEEauthorrefmark}[1]{\raisebox{0pt}[0pt][0pt]{\textsuperscript{\footnotesize#1}}}

\newcommand\rtllmone{RTLLMV1.1\xspace}
\newcommand\rtllmtwo{RTLLMV2\xspace}
\newcommand\verilogeval{VerilogEval\xspace}
\newcommand\verilogevaltwo{VerilogEvalV2\xspace}
\newcommand\codev{CodeV\xspace}

\newcommand\rtlcoder{RTLCoder\xspace}

\newcommand\chipgptv{ChipGPTV\xspace}

%% file: sections/1_introduction.tex
\section{Introduction}

\begin{quote}
\emph{``For better or worse, benchmarks shape a field.''} \\ 
\raggedleft --- David Patterson~\cite{patterson2012better}
\end{quote}

Recently, the application of Large Language Models (LLMs) like GPT~\cite{achiam2023gpt} to automatically generate Verilog code has gained significant attention in the field of hardware design automation~\cite{liu2023chipnemo, liu2023rtlcoder, pei2024betterv, zhang2024mg, zhao2024codev, dataisallyouneed, gao2024autovcoder, liu2024craftrtl, sami2024aivril, chang2023chipgpt, Blocklove_2023, yao2024rtlrewriter, chen2024dawn, yan2024assertllm, fu2023gpt4aigchip, wu2024chateda, kande2023llm, tsai2024rtlfixer, yao2024hdldebugger, yan2023viability}. Based on their powerful code generation capability, LLMs can transform specifications that include natural language descriptions, diagrams, and tables into Verilog code, facilitating the automation of hardware design.

Despite the growing potential of LLMs to assist real-world workflows, existing benchmarks~\cite{liu2024openllm, thakur2023benchmarking, liu2023verilogeval, lu2024rtllm, chang2024natural, pinckney2024revisiting, allam2024rtl, li2024specllm} often fail to accurately capture their true complexity due to three limitations.

\begin{table*}[t]
\caption{Comparison with previous benchmarks, with the averages shown in the table.}
\resizebox{\textwidth}{!}{
\begin{tabular}{|c|c|c|c|c|c|c|c|c|}
\hline
\multirow{2}{*}{\textbf{Benchmarks}}         & \multirow{2}{*}{\textbf{Designs}} & \multicolumn{3}{c|}{\textbf{Design Quality}}       & \multicolumn{2}{c|}{\textbf{Specification Quality}} & \multicolumn{2}{c|}{\textbf{Verification Quality}} \\
\cline{3-9}
                                    &                          & \textbf{Code Lines} & \textbf{Circuit Cells} & \textbf{SubModules} & \textbf{Doc. Lines}          & \textbf{Multi-modal}         & \textbf{Line Coverage}          & \textbf{Formal}          \\
\hline
Thakur et al.                       & 17                       & 16.6        & 7.9            & 0          & 1.8               & \ding{55}                  & 100.0\%                  & \ding{55}              \\
\rtllmone                               & 29                       & 56.1        & 48.1           & 0.3      & 22.3              & \ding{55}                  & 98.5\%                & \ding{55}              \\
\rtllmtwo                              & 50                       & 46.1          & 33.7           & 0.2       & 20.8                & \ding{55}                  & 96.0\%                & \ding{55}              \\
\verilogeval-human                   & 156                      & 15.8         & 31.4           & 0          & 5.7               & \ding{55}                  & 99.7\%                & \ding{55}              \\
\verilogeval-machine                 & 143                      & 13.9        & 6.4             & 0          & 3.2               & \ding{55}                  & 99.7\%                & \ding{55}              \\
\verilogevaltwo                        & 156                      & 16.1        & 31.4           & 0          & 16.5              & \ding{55}                  & 99.7\%                & \ding{55}              \\
\chipgptv                            & 33                       & 51.2        & 18.0           & 0.6      & 12.5              & \checkmark                 & 96.9\%                & \ding{55}
              \\
\hline
\rowcolor[HTML]{EFEFEF}
\xname-module (ours) & 60                       & 241.2        & 0.52K           & 1.6       & 197.3               & \checkmark                 & 100.0\%                  & \checkmark             \\
\rowcolor[HTML]{EFEFEF}
\xname-system (ours) & 4                        & 3537.3       & 1.1M            & 14.5      & 2.9k                & \checkmark                 & 100.0\%                  & \checkmark            \\
\hline
\end{tabular}
}
\label{tab:overview}
\vspace{-10pt}
\end{table*}

\textbf{Existing benchmarks oversimplify task design, while practical projects exhibit complex and hierarchical structures.} Specifically, existing benchmarks are designed for flat, isolated tasks, whereas real-world projects involve multi-level dependencies and dynamic constraints with more module instantiations, more lines of code, a larger number of cells, and hierarchical system structures. 
For example, designs in OpenCores~\cite{opencores}, a set of open-source IPs, contain an average of $\sim180$ lines of Verilog per module while existing benchmarks only contain $\sim50$ lines of code. Additionally, design implementations in the real world often contain more submodule instantiations. 
We will show in our experiments that LLMs perform poorly on Verilog generation tasks that are closer to real-world designs, where even the current best model, o1-preview, achieves only a $13.3\%$ pass@1 on module-level tasks and fails to generate any system-level design.

\textbf{The input format of existing benchmarks largely differs from real-world design specifications and cannot accurately represent the functionality, especially when it comes to complex problems.}
Design specifications in real-world scenarios typically include detailed functional descriptions, diagrams (\eg data flows), and other essential implementation details like interfaces and constraints. In contrast, current benchmarks often rely on simple natural language instructions to describe functions and input/output signals, as well as tables and state transitions in text format (\eg ``a \verb|->| b''), which are insufficient for accurately representing the functionality of the target design.
For example, the real-world design specification of \texttt{SDC\_MMC controller}~\cite{opencores-ds} contains $\sim 800$ lines of text, diagrams, and tables, within $32$ pages, while in existing benchmarks, the average number of lines in LLMs' inputs is only $1.8 \sim 22.3$ and most of them are text-only.

\textbf{The verification environments are not rigorous and cannot meet the requirements of practical usage. }
Current benchmarks utilize predefined testbenches to verify the correctness of the generated Verilog code. However, this approach lacks rigor and can lead to an overestimation of LLMs' accuracy.
For example, we find that testbenches in most existing benchmarks cannot achieve $100\%$ line coverage and \textit{$44.2\%$} of the Verilog code generated by GPT-4-Turbo, which passes the testbenches in \rtllmtwo~\cite{liu2024openllm}, fails in formal verification.

Motivated by such limitations, in this work, we aim to benchmark LLMs' Verilog generation capability in \textit{real-world design workflows} by simulating Verilog coding scenarios that are as close to real-world workflows as possible.
Specifically, we present \textbf{RealBench}, a Verilog generation benchmark that consists of \textit{real-world open-source IP designs} with complex and structured design implementations, multi-modal and formatted design specifications, and rigorous verification environments.

The main features of RealBench include:

\begin{itemize}
    \item \textbf{More complex and structured designs.} Based on four open-source real-world IP cores, including AES encoder/decoder cores, an SD card controller, and a CPU core (Hummingbirdv2 E203~\cite{e203_hbirdv2}), RealBench contains designs with a large number of code lines and a complex, structured design hierarchy, which places higher demands on LLMs' Verilog generation capability. Table~\ref{tab:overview} shows that compared with previous benchmarks, designs in RealBench contain $\sim4.3\times$ code lines, $\sim10.8\times$ circuit cells, and $\sim2.7\times$ submodule instantiations.
    \item \textbf{Multi-modal, detailed, and formatted design specifications.} 
    The specifications encompass detailed functional descriptions, diagrams (\ie block overview, data flows, and state transition), submodules' information (including descriptions and I/O port definitions), and other essential details including interfaces, registers, corner cases, and constraints. We ensure that the design specifications are accurate and formatted by writing them manually. Table~\ref{tab:overview} shows that design specifications in RealBench contain $\sim8.8\times$ longer texts with various multi-modal diagrams. 
    \item \textbf{Rigorous verification process.} We develop testbenches manually to ensure \textit{100\% line coverage} for the reference Verilog. Additionally, we provide a formal verification workflow to verify the equivalence between the LLM-generated code and the reference Verilog.
    \item \textbf{Two-level tasks.} RealBench offers tasks at two distinct levels: module-level and system-level. Module-level tasks assess the capability to generate Verilog modules and instantiate specified submodules. System-level tasks measure the ability to implement the entire system from scratch, given detailed hierarchical design specifications.
\end{itemize}

We conduct extensive evaluations on RealBench with various LLMs of different sizes and types, and two coding agents that can debug their code based on verification feedback. 
The results indicate that: (1) Even o1-preview can only achieve $13.3\%$ formal@1 on module-level tasks and $0\%$ on system-level tasks. (2) The rigorous verification environment is necessary to assess LLMs' performance reliably. 
(3) Reasoning models (\eg DeepSeek-R1-671B) show reduced advantages over the corresponding general models (\eg DeepSeek-V3-671B) in complex tasks. 
(4) The multi-modal LLM, GPT-4o-V, shows a small advantage over the text-only GPT-4o, implying that a stronger multi-modal model is needed to handle design diagrams efficiently. 
(5) A simple debugging agent can improve the correctness of the generated code.

Although current results show that LLMs still struggle in real‑world design workflows, experience from the software‑engineering domain suggests that the development of LLMs is surprisingly fast and that these shortcomings will likely be overcome within years: In just 15 months, the best performance on SWE‑bench~\cite{jimenez2024swebench}, a software‑engineering benchmark, have rocketed from 9.39\% to 23\% (open‑source)~\cite{yang2024swe} and 55\% (closed‑source). Thus, constructing a benchmark that reflects real‑world scenarios has been indispensable for this development, and we hope RealBench will accelerate the rapid progress in hardware design automation.

The contributions of this paper are as follows:
\begin{itemize}
    \item We have identified shortcomings in existing benchmarks with respect to design complexity, input format, and validation rigor, preventing them from effectively benchmarking real‑world Verilog generation. Specifically, we find that existing benchmarks lack rigorous verification processes, with line coverage of only $59.1\%$ in the worst case, and relatively $44.2\%$ of the code passing testbenches fails in formal verification (Fig.~\ref{fig:duijiaoxian}). This may result in an overestimation of the LLMs' performance.
    \item We propose RealBench, the first IP-level Verilog generation benchmark that closely resembles real-world design workflows to facilitate the advancement of hardware design automation. This benchmark includes complex open-source implementations, detailed and formatted design specifications, and rigorous verification environments.
    \item We conduct comprehensive evaluations of various LLMs and agents on RealBench.
    Results show that while these LLMs and the agent perform well on existing benchmarks, there remains a significant gap between their capabilities and their application to real-world workflows.
    \item Our results highlight potential directions in this field: formal verification of LLM-generated code, less hallucinated reasoning models for complex tasks, and stronger models that can handle diagram inputs and submodule instantiations effectively.
\end{itemize}

%% file: sections/2_methods.tex
\section{\xname: towards Real Verilog Generation}
We will introduce \xname in the order of designs, design specifications, verification environments, two difficulty levels of tasks, and our standardized decontamination process.
\subsection{Designs}
\begin{figure}[t]
\centerline{\includegraphics[width=0.95\linewidth]{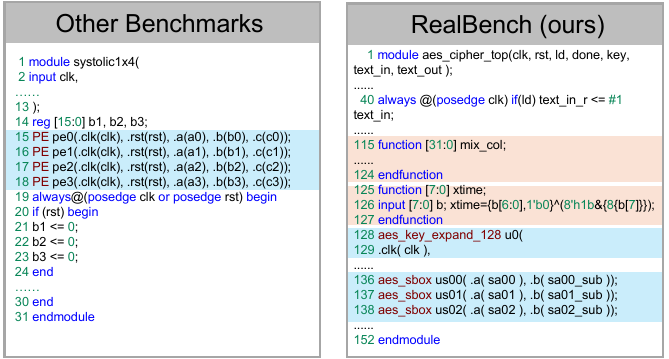}}
\caption{Compared with designs in other benchmarks, ours have more complex module instantiation and are more realistic in reusing functions.}
\label{fig:design}
\vspace{-10pt}
\end{figure}

The comparison between our designs and other benchmarks' is shown in Fig.~\ref{fig:design}.
To ensure the representativeness and correctness of designs in \xname, we build \xname based on a CPU core (Hummingbirdv2 E203~\cite{e203_hbirdv2}) and three IPs from the OpenCores~\cite{opencores} stable stage that are commonly used in SoC design: an SD card controller and AES encoder/decoder cores.
The SD card controller is primarily responsible for complex state and memory management, while the AES encoder/decoder core is mainly focused on computation. In contrast, the CPU core integrates both complex control logic and multiple computational units, making it a particularly challenging benchmark.
The hierarchy of these four designs is shown in Fig.~\ref{fig:callgraph}.

\begin{figure}[t]
\centerline{\includegraphics[width=0.95\linewidth]{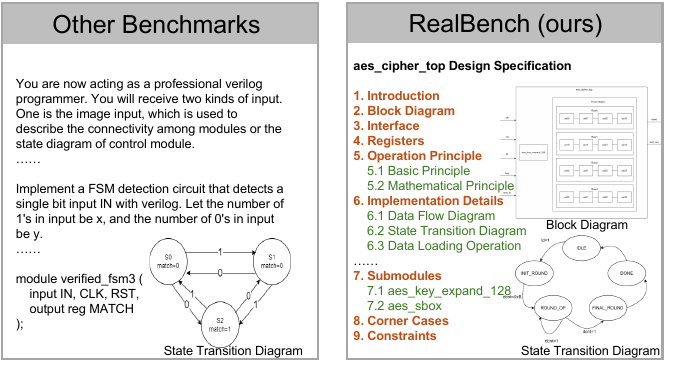}}
\caption{Compared with design specifications in other benchmarks, our design specifications are more detailed and formatted, which aligns with real-world workflows.}
\label{fig:doc}
\vspace{-10pt}
\end{figure}

\begin{figure}[t]
\centerline{\includegraphics[width=0.95\linewidth]{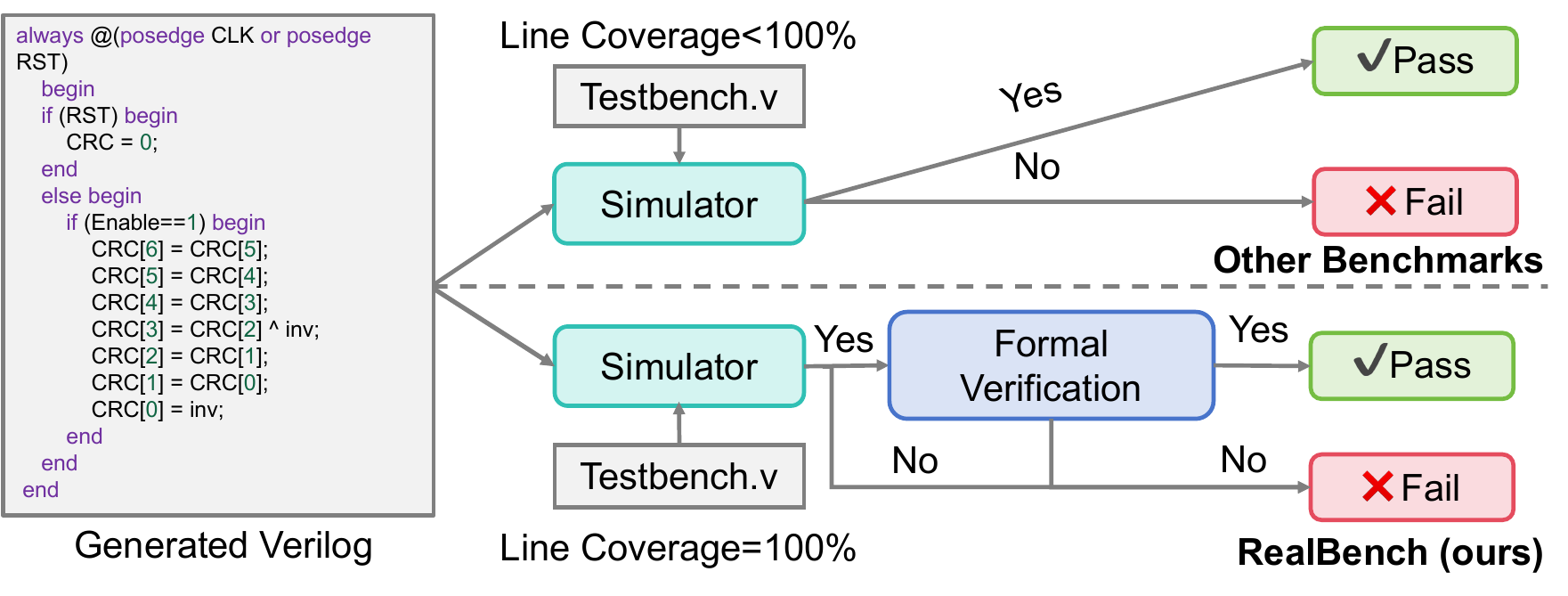}}
\caption{Compared with verification processes in other benchmarks, our verification process has 100\% coverage testbenches and a formal verification process, making results more reliable.}
\label{fig:verification}
\vspace{-10pt}
\end{figure}

\begin{figure}[t]
\centerline{\includegraphics[width=0.6\linewidth]{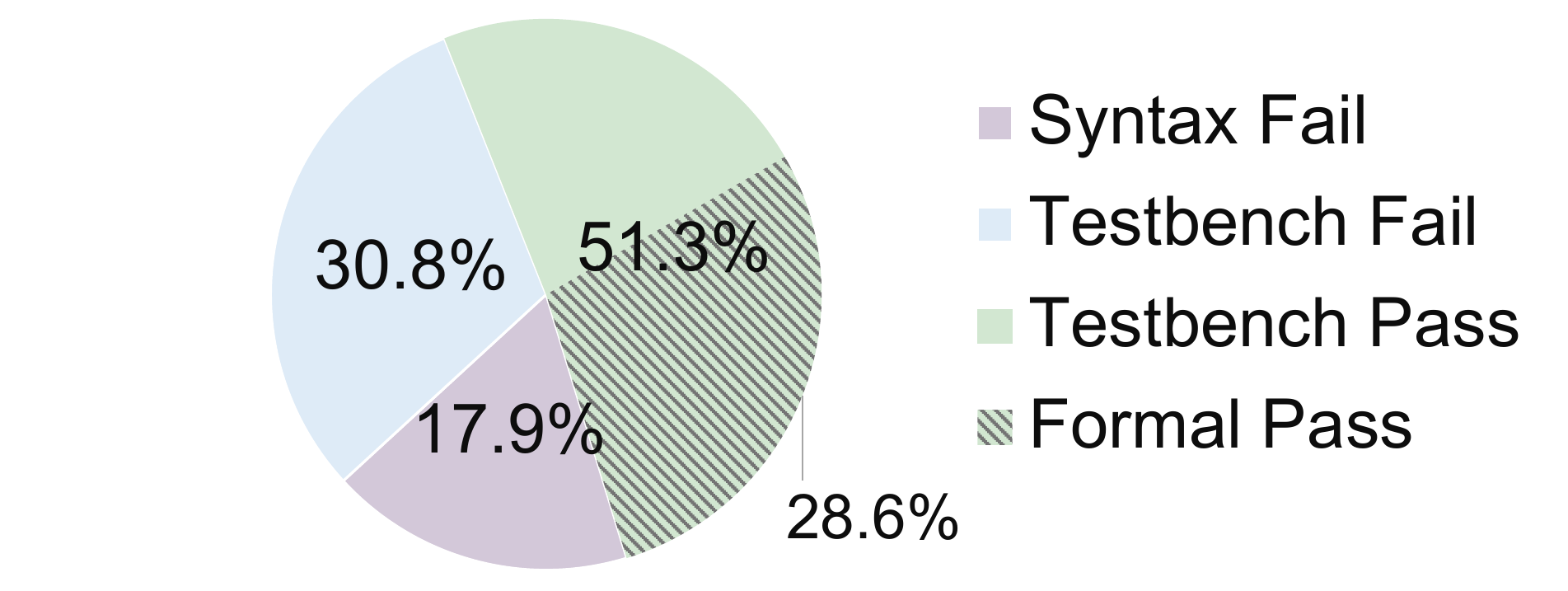}}
\caption{GPT-4-Turbo results on \rtllmtwo. Some code passes the original testbench but fails in our formal verification.}
\label{fig:duijiaoxian}
\vspace{-10pt}
\end{figure}

Table~\ref{tab:details} represents details of all modules in four IPs. We synthesize each module with an open-source EDA tool Yosys~\cite{Yosys} and Nangate45~\cite{Nangate} and report the number of cells.

\subsection{Design Specifications}
Existing open-source design specifications, although well-organized in format, are incomplete in details and thus cannot be used directly to implement their functionality.
Therefore, we rewrite design specifications manually according to the following rules.
The comparison between our design specifications and other benchmarks' is shown in Fig.~\ref{fig:doc}.

\textbf{Principle.} The main principle of handwriting design specifications is to ensure that another engineer can correctly implement the corresponding Verilog code solely based on the given design specifications. 
Specifically, the design specification should 
(1) contain all necessary information for the design and have a consistent format to maintain clarity,
(2) ensure that the function description is straightforward and unambiguous,
and (3) use intuitive diagrams and tables to illustrate key concepts like interfaces.

\textbf{Hierarchical structure.}
We structure design specifications in a hierarchical structure which keeps the same structure as the code implementation. 
That is, (1) each module of the design has its own distinct design specification, in which we detail the interface and functional information of all submodules.
(2) Cross-level information is not permitted in design specifications. For example, diagrams of a module should not contain unnecessary details of its submodules.

\textbf{Co-design with the testbench verification.}
We use specification and testbench verification co-design to better describe the function details.
Specifically, while developing testbenches by manually analyzing code and inserting key functional points, we reflect these functional points in the design specification to ensure the specification describes key functions clearly.

\textbf{Format and content.}
We keep the format and content of design specifications similar to that of open-source industry standards~\cite{opencores-ds} to assess LLMs' Verilog generation ability when working in real-world design workflows.
\begin{itemize}
    \item Module introduction. We briefly describe the overall function and provide an overview diagram.
    \item I/O interface. This is a table containing information such as port name, direction, width, and description, which describes the format of the module's inputs and outputs.
    \item Clock and reset. This includes descriptions of the module's clock signal and reset signal, such as whether the reset signal is asynchronous.
    \item Submodule information. If the current module instantiates other submodules, the information of submodules (\eg description, I/O interface, etc.) is provided in the submodule information.
    \item Detailed function description. This includes state transitions and executed operations. State transitions are provided through diagrams and accompanying textual descriptions, while executed operations are explained through key execution processes.
    \item Additional information. Additional information includes registers (a table containing register names, width, and description), corner cases, constraints, etc.
\end{itemize}

Additionally, we ensure the accuracy, consistency, and standardization of design specifications through multiple rounds of review and revisions by different writers.

\subsection{Verification Environments}
\xname verification workflow is shown in Fig.~\ref{fig:verification}.

\textbf{Testbench verification.}
The design of our testbench was inspired by \verilogeval~\cite{liu2023verilogeval}, where the main component of the testbench is an input generator. 
Through a more detailed manual analysis of the design's code implementation and the insertion of functional points, we increased the testbench's line coverage of all designs to $100\%$.
This makes our verification results more reliable compared with existing benchmarks.

\textbf{Formal verification.}
A testbench achieving $100\%$ line coverage of the reference code provides more rigorous verification than current benchmarks, but still cannot guarantee that the LLM-generated code is completely correct.
Therefore, we provide a formal verification workflow.
First, we use an open-source EDA tool Yosys~\cite{Yosys} to synthesize both the reference and generated Verilog code into netlists, then perform formal equivalence checking on the two netlists using Cadence JasperGold~\cite{jaspergold}.

\subsection{Module-Level and System-Level Tasks}
\xname provides two levels of task difficulty: module-level and system-level. Each level includes different design specifications as LLMs' input and verification environments for validating results.

\textbf{Module-level tasks.} 
Module-level tasks evaluate the ability of LLMs to generate individual modules.
For each module-level task, the input provided to the LLM includes the design specification corresponding to that module. 
If the target module needs to instantiate other submodules (\ie it is not a leaf node in the design hierarchy), the input will also provide submodules' information. 
During verification, we use golden submodule implementations to ensure the submodules' correctness so that LLMs are not required to implement them. 
This aligns with real-world workflows, as module-level implementation and verification do not need to consider the correctness of other modules.

\textbf{System-level tasks.}
System-level tasks evaluate LLMs' ability to generate a complete design from scratch. The input for each system-level task consists of the design specifications for the entire system, which include specifications at each level of the design hierarchy. No additional information is provided beyond these design specifications. Therefore, if a module requires instantiating other submodules, the LLM must implement those submodules. During verification, we focus on validating the system-level testbench, without delving into the implementation details of individual modules, as variations in module details may not affect the overall system correctness.

\subsection{Standardized Decontamination Process}
The existing benchmarks do not provide a standardized decontamination process, which makes comparisons between different methods potentially unfair. 
For example, \rtlcoder~\cite{liu2023rtlcoder} and \codev~\cite{zhao2024codev} decontaminate their training data with RougeL, while many others do not explicitly mention their decontamination process.
To ensure a fair evaluation, we provide a standardized decontamination process for LLM fine-tuning methods to remove training data similar to \xname. 
Following \rtlcoder and \codev, we use RougeL~\cite{lin2004rouge}, a widely used similarity metric, to measure the similarity between data and suggest setting $\beta=1.0$ and removing training data whose RougeL is greater than $0.5$.

\begin{table*}[!htbp]
\caption{Module details of the four IPs in \xname.}
\centering
\begin{tabular}{|>{\raggedright\arraybackslash}m{2cm}|>{\raggedright\arraybackslash}m{2.3cm}|>{\raggedright\arraybackslash}p{9.5cm}|>{\centering\arraybackslash}m{0.5cm}|>{\centering\arraybackslash}m{0.5cm}|>{\centering\arraybackslash}m{0.7cm}|}

\hline 
\textbf{Design}  & \textbf{Module Name} & \textbf{Description} & \textbf{Code Lines} &\textbf{Doc. Lines}& \textbf{Circuit Cells} \\
\hline
\multirow{14}{*}{sdc\_controller}                       & sd\_bd                  & Managing the buffer descriptors used for data transmission and reception between the system memory and the SD card.                                                                                                                                                                                                                         & 141                           & 51                           & 1739                     \\
                                                        & sd\_clock\_divider      & A converter between the system clock and the clock required by the SD card-related modules.                                                                                                                                                                                                                                             & 35                            & 32                           & 50                       \\
                                                        & sd\_crc\_16             & Computing a 16-bit CRC value based on the input data bits.                                                                                                                                                                                                                                                                                  & 34                            & 31                           & 55                       \\
                                                        & sd\_crc\_7              & Designed to compute a 7-bit CRC value for serial data inputs.                                                                                                                                                                                                                                                 & 26                            & 43                           & 24                       \\
                                                        & sd\_controller\_wb      & The wishbone slave interface, which manages the reading and writing to the configuration registers.                                                                                                                                                                                                                                     & 307                           & 279                          & 923                      \\
                                                        & sd\_data\_master        & Managing data transfer between the SD card and the host system.                                                                                                                                                                                                                                                                             & 404                           & 256                          & 589                      \\
                                                        & sd\_cmd\_master         & Synchronizing the communication from the host interface with the physical interface.                                                                                                                                                                                                                                                        & 245                           & 353                          & 545                      \\
                                                        & sd\_rx\_fifo            & Buffering incoming data from the SD card before it's read by the host system.                                                                                                                                                                                                                                           & 106                           & 73                           & 1143                     \\
                                                        & sd\_tx\_fifo            & A dual-clock domain FIFO designed to transfer data between two potentially asynchronous clock domains.                                                                                                                                                                                                                                  & 51                            & 99                           & 1017                     \\
                                                        & sd\_fifo\_rx\_filler    & Managing the reception of data from the SD card and transfers it to the system memory via a Wishbone bus interface.                                                                                                                                                                                                                         & 82                            & 107                          & 241                      \\
                                                        & sd\_fifo\_tx\_filler    & \begin{tabular}[c]{@{}l@{}}Managing the transmission FIFO buffer for the data stream.\end{tabular}                                                                                                                                                                                                                                        & 92                            & 98                           & 241                      \\
                                                        & sd\_data\_serial\_host  & The interface towards physical SD card device data port.                                                                                                                                                                                                                                                                                & 374                           & 252                          & 1031                     \\
                                                        & sd\_cmd\_serial\_host   & An interface to the external SD /MMC card.                                                                                                                                                                                                                           & 468                           & 358                          & 1525                     \\
                                                        & sdc\_controller         & The top module of sdc\_controller.                                                                                                                                                                                                                                                                                                      & 388                           & 362                          & 238                      \\
\hline
\multirow{2}{*}{aes\_inv\_cipher\_top}                  & aes\_inv\_sbox          & Performing the inverse non-linear byte substitution operation.                                                                                                                                                                                                                                                                               & 264                           & 178                          & 453                      \\
                                                        & aes\_inv\_cipher\_top   & The core control module of the AES decryption system.                                                                                                                                                                                                       & 210                           & 267                          & 7322                     \\
\hline
aes\_cipher\_top                                        & aes\_cipher\_top        & The core control module of the entire AES encryption system.                                                                                                                                                                                                & 152                           & 182                          & 1445                     \\
\hline
\multirow{3}{*}{\parbox{2cm}{aes\_cipher\_top, aes\_inv\_cipher\_top}} & aes\_key\_expand\_128   & This module expands a 128-bit initial key into round keys.                                                                                                                                                                                                                                                              & 24                            & 121                          & 569                      \\
                                                        & aes\_sbox               & Performing non-linear byte substitution operations.                                                                                                                                                                                                                                                                                          & 264                           & 65                           & 458                      \\
                                                        & aes\_rcon               & Generating constant values used in the AES key expansion process.                                                                                                                                                                                                                                                                           & 31                            & 82                           & 44                       \\
\hline
\multirow{40}{*}{e203\_cpu}                             & e203\_biu               & Managing bus requests between the processor's internal units and the external memory system.                                                                                                                                                                  & 757                           & 317                          & 84                       \\
                                                        & e203\_clk\_ctrl         & Managing clock gating for the processor's core and memory modules.                                                                                                                                                                                                                      & 97                            & 102                          & 18                       \\
                                                        & e203\_clkgate           & Managing clock propagation through enable signals.                                                                                                                                                                                                     & 30                            & 54                           & 3                        \\
                                                        & e203\_core              & The core of the E203 processor.                                                                                                                     & 704                           & 560                          & 4                        \\
                                                        & e203\_cpu               & The top-level module of a RISC-V processor.                                                                                                                                       & 643                           & 593                          & 7                        \\
                                                        & e203\_cpu\_top          & The top-level module in the E203 series.                                                          & 418                           & 249                          & 2                        \\
                                                        & e203\_dtcm\_ctrl        & The DTCM controller in the E203 processor.                                                                                                                                             & 255                           & 279                          & 3                        \\
                                                        & e203\_dtcm\_ram         & The DTCM RAM module for the E203 processor.                                                                                                                                                       & 48                            & 82                           & 1                        \\
                                                        & e203\_extend\_csr       & A stub module that provides interface definitions without actual functionality.                                                                                                                                                        & 28                            & 27                           & 0    \\
                                                        & e203\_exu               & The execution unit of the processor core.                                      & 724                           & 705                          & 15                       \\
                                                        & e203\_exu\_alu          & A core component of the E203 processor's execution unit.                                                                           & 682                           & 456                          & 935                      \\
                                                        & e203\_exu\_alu\_bjp     & Conditional branch and jump instructions.         & 77                            & 86                           & 104                      \\
                                                        & e203\_exu\_alu\_csrctrl & Controlling CSR read/write operations.                                                                                                                                                                                                     & 88                            & 89                           & 153                      \\
                                                        & e203\_exu\_alu\_dpath   & The datapath implementation of the ALU in the E203 processor.                                                                                                                          & 378                           & 87                           & 1465                     \\
                                                        & e203\_exu\_alu\_lsuagu  & Implementing the AGU for load/store and atomic memory operation instructions.                                                                                                                                                        & 461                           & 176                          & 544                      \\
                                                        & e203\_exu\_alu\_muldiv  & Implementing a 17-cycle multiplier and a 33-cycle divider.                                                   & 45                            & 136                          & 1203                     \\
                                                        & e203\_exu\_alu\_rglr    & Implementing regular ALU instructions.                                                                                                                                                                                                                                                                                             & 72                            & 75                           & 73                       \\
                                                        & e203\_exu\_branchslv    & Implementing branch resolution in the RISC-V processor.                                                                                                            & 79                            & 75                           & 595                      \\
                                                        & e203\_exu\_commit       & Implementing the commit stage of the core pipeline.                                                                                                                                 & 247                           & 210                          & 116                      \\
                                                        & e203\_exu\_csr          & Implementing the core's CSR interface.                                                                                              & 428                           & 255                          & 1083                     \\
                                                        & e203\_exu\_decode       & This module decodes 32-bit and 16-bit instructions.                                                                                                                                                                         & 907                           & 384                          & 587                      \\
                                                        & e203\_exu\_disp         & Dispatching instructions to functional units like the ALU and OITF.                                                                                      & 140                           & 139                          & 149                      \\
                                                        & e203\_exu\_excp         & Handling exceptions and interrupts in the RISC-V processor.                                                                                                                    & 301                           & 221                          & 351                      \\
                                                        & e203\_exu\_longpwbck    & Implementing the Write-Back (WB) logic.                                                                                                      & 135                           & 83                           & 126                      \\
                                                        & e203\_exu\_nice         & The bridge between the execution unit and custom hardware accelerators.                                                                                                                 & 75                            & 108                          & 12                       \\
                                                        & e203\_exu\_oitf         & Managing the lifecycle of long-pipeline instructions.                                                                                              & 118                           & 65                           & 168                      \\
                                                        & e203\_exu\_regfile      & Implementing the integer general-purpose register file.                                                                                             & 63                            & 59                           & 1780                     \\
                                                        & e203\_exu\_wbck         & Arbitrating write-back requests.                                                                                                                                                  & 59                            & 41                           & 44                       \\
                                                        & e203\_ifu               & The top-level module of the E203 processor's instruction fetch unit.                                 & 168                           & 271                          & 2                        \\
                                                        & e203\_ifu\_ifetch       & The core of the E203 processor's instruction fetch unit.                        & 321                           & 589                          & 522                      \\
                                                        & e203\_ifu\_ift2icb      & A core component of the E203 processor's instruction fetch unit.                                                                                                  & 414                           & 340                          & 459                      \\
                                                        & e203\_ifu\_litebpu      & The branch prediction unit of the E203 processor.                                                                            & 57                            & 179                          & 166                      \\
                                                        & e203\_ifu\_minidec      & A mini decoder in the E203 processor's IFU.                                                                         & 75                            & 118                          & 1                        \\
                                                        & e203\_irq\_sync         & An interrupt synchronization module in the E203 processor,.                                                                                                                                           & 83                            & 107                          & 4                        \\
                                                        & e203\_itcm\_ctrl        & A multi-port memory access controller that manages and arbitrates access requests to the ITCM from the IFU, LSU, and external interfaces.                                                                                                  & 419                           & 325                          & 305                      \\
                                                        & e203\_itcm\_ram         & The ITCM RAM module for the E203 processor.                                                                                                                                                & 50                            & 91                           & 1                        \\
                                                        & e203\_lsu               & The memory access unit of the E203 processor. & 254                           & 252                          & 2                        \\
                                                        & e203\_lsu\_ctrl         & The LSU control module in the E203 processor.                                                   & 733                           & 423                          & 476                      \\
                                                        & e203\_reset\_ctrl       & Managing reset signals for critical system modules.                                                                                                                                                                                                                 & 65                            & 49                           & 3                        \\
                                                        & e203\_srams             & The memory management module of the E203 processor.                                                                                                                             & 77                            & 94                           & 2                       \\
\hline
\end{tabular}
\label{tab:details}
\vspace{-10pt}
\end{table*}

%% file: sections/3_experiments.tex
\section{Evaluations}

In this section, we evaluate the performance of LLMs and LLM-based agents on \xname and existing benchmarks, investigate whether LLMs can accomplish real-world design tasks, and provide possible future directions for hardware design automation through a detailed analysis of the results.
\subsection{Evaluation Setup}
\textbf{Other benchmarks.}
We compare our benchmark with the most recent benchmarks, including \rtllmtwo~\cite{liu2024openllm}, \verilogevaltwo~\cite{pinckney2024revisiting}, and \chipgptv~\cite{chang2024natural}.

\begin{figure}[t]
\centerline{\includegraphics[width=0.9\linewidth]{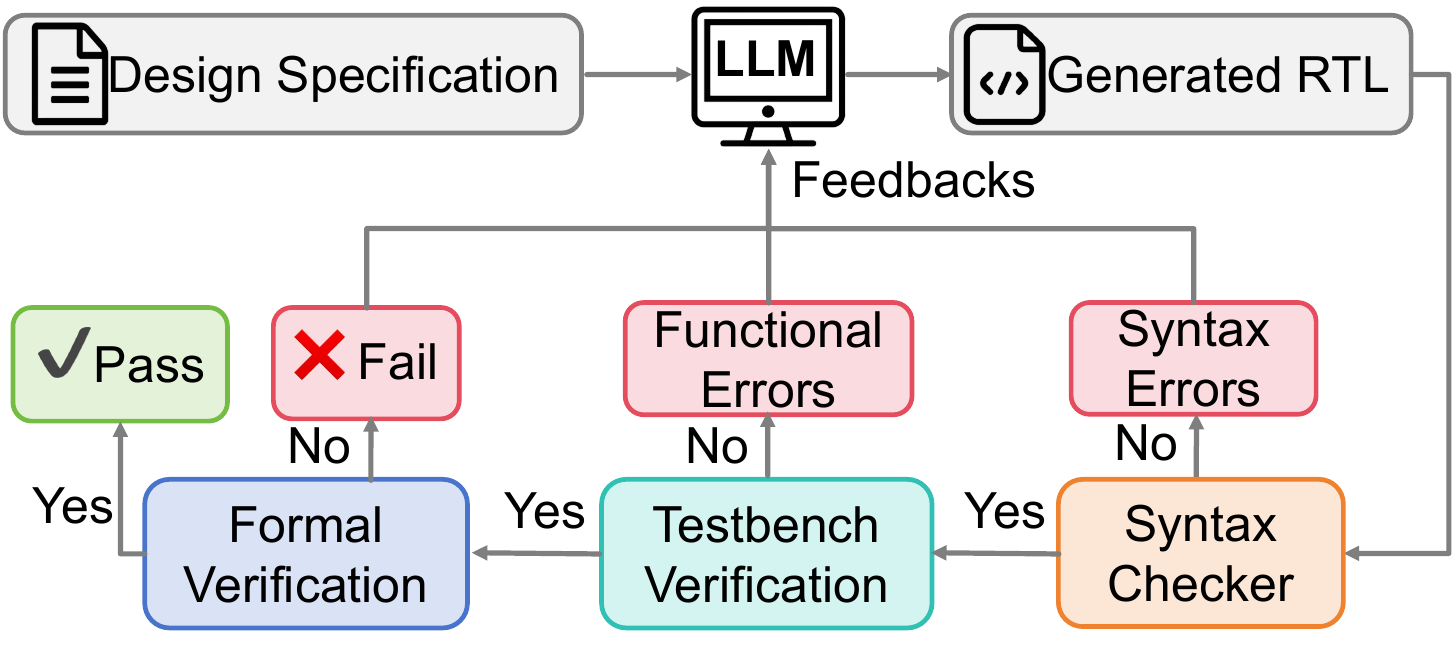}}
\caption{Self-reflection agent overview. The agent generates Verilog code, receives feedback from the syntax checker, testbench verification and formal verification, and then modifies the code based on the feedback.}
\label{fig:agent-overview}
\vspace{-10pt}
\end{figure}

\textbf{Models.}
We comprehensively tested the performance of most advanced models, including the GPT series~\cite{achiam2023gpt}, LLama-3.1~\cite{dubey2024llama}, DeepSeek-V3~\cite{deepseekv3}, DeepSeek-R1~\cite{deepseekr1}, the best model of RTLCoder~\cite{liu2023rtlcoder}, and the best model of CodeV~\cite{zhao2024codev}. 
These models encompass different sizes, open-source and closed-source, and single-modal and multi-modal types.
We list the model abbreviations and their corresponding full names as follows:
\begin{itemize}
    \item DS-V3-671B: The 671B DeepSeek-V3 model~\cite{deepseekv3}.
    \item DS-R1-671B: The 671B DeepSeek-R3 model~\cite{deepseekr1}.
    \item RTLCoder-DS-6.7B: RTLCoder~\cite{liu2023rtlcoder} trained on 6.7B DeepSeek-Coder~\cite{deepseekcoder}, which is one of its best models.
    \item CodeV-QW-7B: CodeV~\cite{zhao2024codev} trained on 7B CodeQwen~\cite{codeqwen}, which is one of its best models.
    \item GPT-4o: OpenAI GPT-4o \textit{w/o} vision input (text-only).
    \item GPT-4o-V: OpenAI GPT-4o \textit{w/} vision input.
\end{itemize}

\begin{table*}[ht]
    \centering
    \caption{The main results of various models across different benchmarks.}
    \begin{tabular}{|>{\raggedright\arraybackslash}m{1.5cm}|>{\raggedright\arraybackslash}m{1cm}|>{\centering\arraybackslash}m{0.9cm}|>{\centering\arraybackslash}m{0.9cm}|>{\centering\arraybackslash}m{0.9cm}|>{\centering\arraybackslash}m{1.2cm}|>{\centering\arraybackslash}m{0.9cm}|>{\centering\arraybackslash}m{0.9cm}|>{\centering\arraybackslash}m{1.2cm}|>{\centering\arraybackslash}m{0.9cm}|>{\centering\arraybackslash}m{0.9cm}|>{\centering\arraybackslash}m{0.9cm}|}
    \hline 
    \textbf{Benchmarks}   & \textbf{Metrics}   & \textbf{GPT-4-Turbo} & \textbf{o1-preview}& \textbf{Llama-3.1-8B} & \textbf{Llama-3.1-405B} & \textbf{DS-V3-671B} & \textbf{DS-R1-671B} & \textbf{\rtlcoder-DS-6.7B} & \textbf{CodeV-QW-7B} & \textbf{GPT-4o} & \textbf{GPT-4o-V} \\
    \hline 

\multirow{4}{*}{\rtllmtwo}                            & syntax@1 & 82.1        & 94.0       & 92.4         & 75.7           & 89.9       & 90.3       & 79.8          & 86.6     & 92.4   & -                    \\
                                                     & syntax@5 & 88.9        & -          & 95.9         & 84.2           & 94.4       & 95.6       & 87.2          & 94.2     & 95.9   & -                    \\
                                                     & func@1   & 51.3        & 64.0       & 62.8         & 45.0           & 60.1       & 65.4       & 45.7          & 43.6     & 62.8   & -                    \\
                                                     & func@5   & 62.4        & -          & 71.9         & 53.3           & 68.0       & 75.6       & 51.0          & 58.1     & 71.9   & -                    \\
\hline 

\multirow{4}{*}{\parbox{1.5cm}{\verilogevaltwo-spec2rtl}}            & syntax@1 & 98.5        & 98.7       & 89.8         & 98.0           & 99.9       & 99.5       & 97.5          & 84.0     & 100.0  & -                    \\
                                                     & syntax@5 & 99.6        & -          & 98.7         & 98.7           & 100.0      & 100.0      & 99.7          & 97.3     & 100.0  & -                    \\
                                                     & func@1   & 62.0        & 70.5       & 24.4         & 56.6           & 56.5       & 75.4       & 35.1          & 39.7     & 64.6   & -                    \\
                                                     & func@5   & 65.6        & -          & 32.1         & 60.7           & 63.0       & 84.1       & 41.4          & 56.4     & 68.2   & -                    \\
\hline

\multirow{4}{*}{\parbox{1.5cm}{\verilogevaltwo-code-complete}}        & syntax@1 & 98.7        & 100.0      & 80.1         & 98.7           & 96.9       & 98.1       & 97.0          & 97.4     & 99.7   & -                    \\
                                                     & syntax@5 & 99.8        & -          & 92.6         & 99.3           & 98.8       & 99.9       & 99.5          & 99.9     & 100.0  & -                    \\
                                                     & func@1   & 41.1        & 71.8       & 6.3          & 56.1           & 62.3       & 76.3       & 39.0          & 50.5     & 62.6   & -                    \\
                                                     & func@5   & 49.8        & -          & 11.0         & 61.8           & 68.7       & 83.0       & 45.7          & 58.8     & 66.9   & -                    \\
\hline

\multirow{4}{*}{\parbox{1.5cm}{\chipgptv}}                            & syntax@1 & 54.6        & 57.6       & 20.8         & 48.6           & 53.0       & 59.7       & 40.2          & 19.4     & 55.9   & 56.0                 \\
                                                     & syntax@5 & 60.4        & -          & 44.0         & 58.6           & 57.0       & 63.6       & 51.8          & 29.1     & 60.6   & 60.3                 \\
                                                     & func@1   & 31.5        & 39.4       & 8.6          & 22.3           & 31.1       & 39.4       & 17.0          & 6.8      & 34.9   & 40.9                 \\
                                                     & func@5   & 39.9        & -          & 20.1         & 26.1           & 36.9       & 48.0       & 23.1          & 13.3     & 41.5   & 47.3                 \\
\hline

\rowcolor[HTML]{EFEFEF}
& syntax@1 & 27.0        & 28.3       & 10.3         & 6.6            & 24.9       & 13.4       & 10.9          & 3.6      & 24.5   & 29.3                 \\
\rowcolor[HTML]{EFEFEF}
& syntax@5 & 36.2        & -          & 21.7         & 15.7           & 32.0       & 30.4       & 19.0          & 7.4      & 29.7   & 38.1                 \\
\rowcolor[HTML]{EFEFEF}
& func@1   & 9.4         & 13.3       & 2.9          & 2.8            & 12.8       & 8.0        & 3.4           & 0.3      & 10.9   & 13.5                 \\
\rowcolor[HTML]{EFEFEF}
& func@5   & 13.8        & -          & 6.2          & 7.6            & 17.3       & 16.2       & 8.5           & 1.0      & 14.0   & 16.2                 \\
\rowcolor[HTML]{EFEFEF}
& formal@1 & 4.4         & 13.3       & 2.6          & 2.6            & 10.8       & 7.8        & 2.0           & 0.0      & 8.8    & 11.7                 \\
\rowcolor[HTML]{EFEFEF}
\multirow{-6}{*}{\parbox{1.5cm}{\xname-Module (ours)}}& formal@5 & 6.8         & -          & 4.7          & 6.9            & 14.2       & 15.4       & 3.6           & 0.0      & 12.4   & 13.3                 \\
\hline

\rowcolor[HTML]{EFEFEF}
& syntax@1 & 0.0         & 25.0       & 0.0          & 0.0            & 41.3       & 16.3       & 0.0           & 0.0      & 31.3   & 38.8                 \\
\rowcolor[HTML]{EFEFEF}
& syntax@5 & 0.0         & -          & 0.0          & 0.0            & 67.6       & 42.2       & 0.0           & 0.0      & 49.3   & 70.1                 \\
\cline{2-12}
\multirow{-3}{*}{\parbox{1.5cm}{\cellcolor[HTML]{EFEFEF} \xname-System (ours)}}&\multicolumn{11}{c|}{\cellcolor[HTML]{EFEFEF} func@1, func@5, formal@1, formal@5 are all 0.}\\
\hline
\end{tabular}
\label{tab:main results}
\vspace{-10pt}
\end{table*}

\textbf{\begin{figure*}[ht]
\centerline{\includegraphics[width=\linewidth]{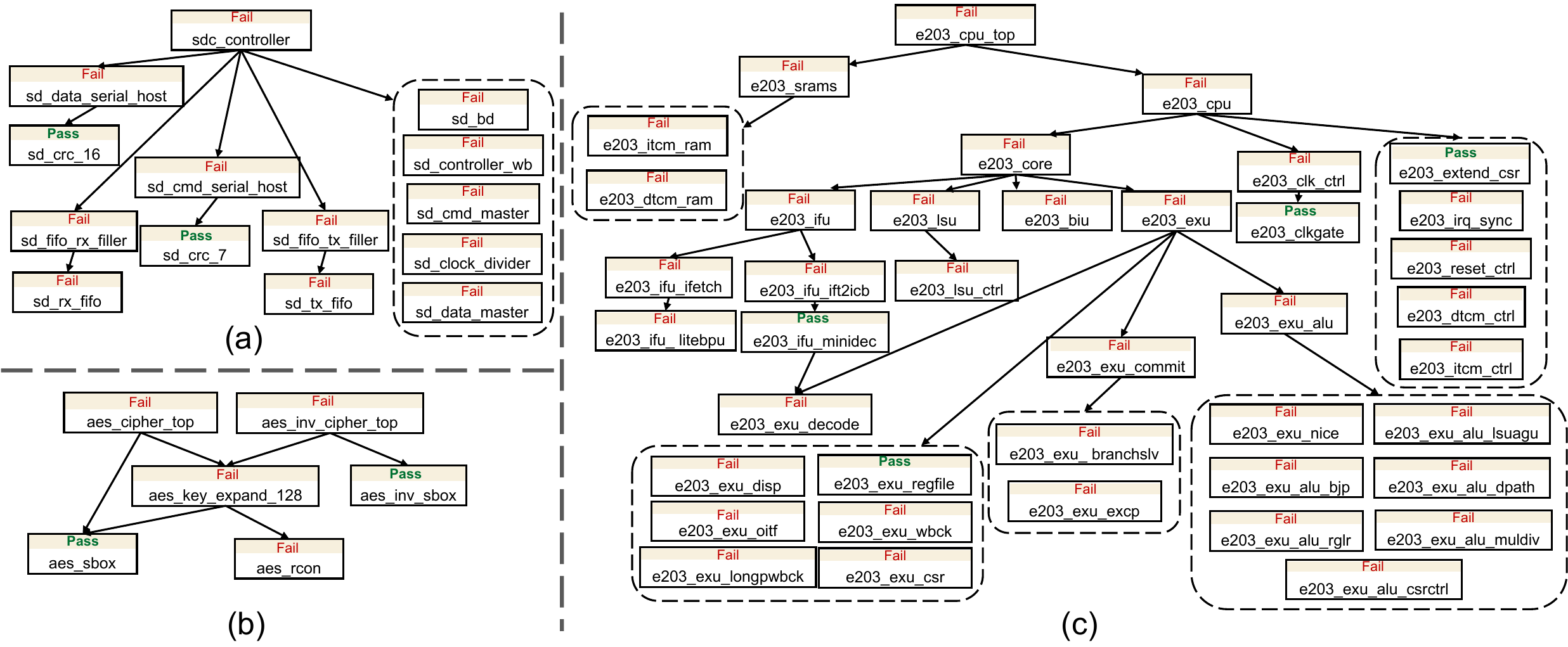}}
\caption{Design hierarchy and o1-preview's formal@1 of (a) SD card controller, (b) AES encoder/decoder core, and (c) Hummingbirdv2 E203 CPU Core. ``Pass'' means o1-preview generates the module correctly while ``Fail'' means the opposite.}
\label{fig:callgraph}
\vspace{-10pt}
\end{figure*}}

\begin{table}[t]
\centering
\caption{The comparison among tasks on submodules and circuit types.}
\setlength{\tabcolsep}{1pt}
\begin{tabular}{|>{\raggedright\arraybackslash}m{1.4cm}|>{\centering\arraybackslash}m{1.6cm}|>{\centering\arraybackslash}m{0.8cm}|>{\centering\arraybackslash}m{0.9cm}|>{\centering\arraybackslash}m{0.6cm}|>{\centering\arraybackslash}m{0.6cm}|>{\centering\arraybackslash}m{0.6cm}|>{\centering\arraybackslash}m{0.6cm}|}
\hline
\textbf{Submodule}              & \textbf{Circuit Type} & \textbf{GPT-4-Turbo} & \textbf{o1-preview} & \textbf{DS-V3-671B} & \textbf{DS-R1-671B} & \textbf{GPT-4o} & \textbf{GPT-4o-V} \\
\hline
\multirow{3}{*}{\parbox{1.4cm}{w/o submodules}} & Comb                  & 0.8                 & 33.3               & 29.6               & 10.4               & 20.4           & 24.6             \\
                                & Seq w/o FSM           & 1.3                 & 50.0                  & 42.5                & 28.8               & 42.5            & 57.5              \\
                                & Seq w FSM             & 0.0                    & 0.0                   & 2.9                & 7.9                & 2.9            & 0.0                 \\
\hline
\multirow{3}{*}{\parbox{1.4cm}{w/ \\submodules}}  &Comb                  & 8.6                 & 6.9                 & 3.6                & 5.9                & 3.3            & 6.0              \\
                                & Seq w/o FSM           & 0.0                    & 0.0                   & 0.0                   & 0.0                   & 0.0               & 0.0                 \\
                                & Seq w FSM             & 0.0                    & 0.0                   & 0.0                   & 0.7                & 0.0               & 0.0                \\
\hline
\end{tabular}
\label{tab:submodule}
\vspace{-10pt}
\end{table}

\textbf{Agents.}
We evaluate the effectiveness of LLM single and multi-agent frameworks on our RealBench.
For the single-agent framework, we develop a self-reflection agent that can generate Verilog based on design specifications, receive feedback from the syntax, testbench, and formal verification, and modify the generated code according to the feedback (Fig.~\ref{fig:agent-overview}).
For the multi-agent framework, we use the SOTA framework VerilogCoder~\cite{verilogcoder}, which firstly generates a plan, executes the plan to generate code, and then debugs the code based on feedback, with each process utilizing different agents. It provides fine-grained feedback to agents through tools including a syntax checker, a simulator, and an AST-based waveform tracing tool.
To fairly compare the performance of two agent frameworks, we use GPT-4o 
as the base LLM, set a 4-hour timeout limit, and only retain the first 6 Verilog code generations to evaluate their generation efficiency. For VerilogCoder, all additional settings remain the same as their original ones.

\textbf{Evaluation metrics.}
Our evaluations mainly focus on three aspects: syntax correctness, function correctness (\ie testbench), and formal correctness. These metrics are measured using metric@1 and metric@5, which assess the correctness of the generated Verilog in one attempt and five attempts, respectively.

\subsection{Results on Single LLMs}
The main results are shown in Table~\ref{tab:main results}, where we set temperatures to 0.2 and sample 20 times to estimate metric@1 and metric@5 except for o1-preview, which we sample only once due to its high cost. 
We can conclude that:

\textbf{Formal verification is necessary.}
Although \xname's testbenches have achieved $100\%$ line coverage, there remains a gap of $0\%$ to $100\%$ relatively between func@1 and formal@1, highlighting that only formal verification can ensure the reliability of the results.
This is also indicated by Fig.~\ref{fig:duijiaoxian}, where we test GPT-4-Turbo on \rtllmtwo with both the original testbenches and our formal verification, and find that relatively $44.2\%$ of the code passing testbenches fails in formal verification.
However, formal verification is slow for large designs. 
Therefore, exploring more efficient methods to apply formal verification to long and unstable code generated by LLMs might be important for future research.

\textbf{All models perform poorly when faced with real-world design requirements, especially for modules that require instantiating submodules.} None of the models succeeded in completing any system-level tasks. Even the best-performing model, o1-preview, only achieved $13.3\%$ formal@1 accuracy in module-level tasks. A more detailed analysis of o1-preview's correctness on the design is shown in Fig.~\ref{fig:callgraph}, which reveals that leaf nodes in the design hierarchy tend to have higher generation accuracy, whereas modules requiring submodule instantiations are generally generated incorrectly.

\textbf{When facing complex Verilog generation tasks, the advantage of reasoning models (\eg DS-R1-671B) over general models (\eg DS-V3-671B) diminishes.} Specifically, on \rtllmtwo, \verilogevaltwo, and \chipgptv, DS-R1-671B improves by 5.3\%, 18.9\%, 14.0\%, and 8.3\% relative to DS-V3-671B, but there is a 3.0\% decrease on \xname formal@1.
A detailed analysis shows that this is due to DS-R1-671B exhibiting more severe hallucination behaviors when facing complex tasks, resulting in simple mistakes, such as generating fictitious module headers. This provides an empirical reference for the development and evaluation of future reasoning models.

\textbf{Existing multi-modal models struggle to effectively handle multimodal information in real-world design specifications}. On the \chipgptv benchmark, we find that the multi-modal model outperforms text-only models, where GPT-4o-V demonstrates a 6.0\% improvement over GPT-4o relative to func@1. 
However, on \xname, the improvement of the multi-modal model is smaller, with only 2.6\% relative to func@1 and 2.9\% relative to formal@1. The reason may be that each task in \chipgptv involves only a single diagram, while tasks in \xname often include multiple diagrams, such as block overviews, data flows, and state transitions, making the processing of vision information harder.
Furthermore, more complex tasks with longer design specifications and code can also lead to a decrease of improvement.

\begin{figure*}[t]
\centerline{\includegraphics[width=\linewidth]{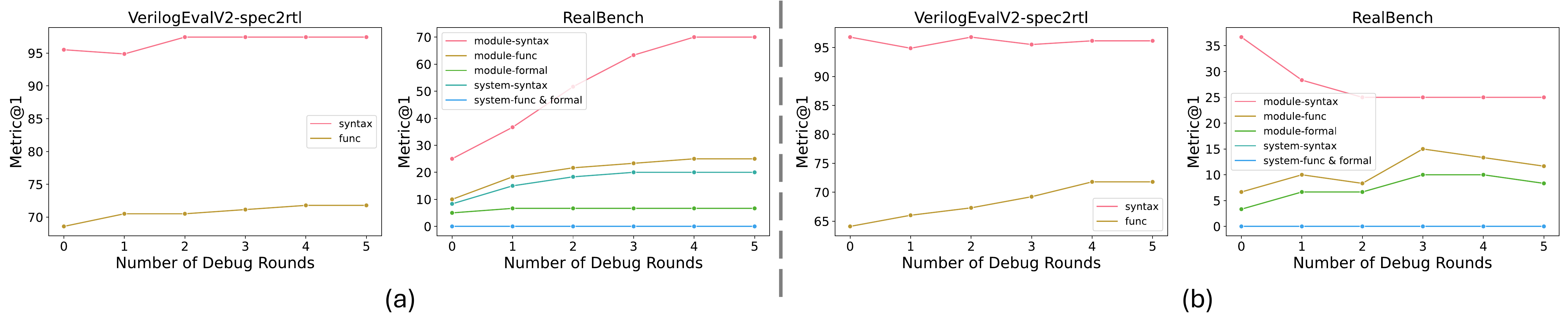}}
\caption{Results of two LLM agent frameworks on \verilogevaltwo-spec2rtl and \xname: (a) Self-reflection agent and (b) VerilogCoder.}
\label{fig:code-agent-result}
\vspace{-10pt}
\end{figure*}

\subsection{Results on Agents}
\textbf{Simple debugging on \xname can also improve the pass rate.} As shown in Fig.~\ref{fig:code-agent-result} (a), although the system-level pass rate remains at 0, the pass rate at the module level continues to improve as the iterations progress.

\textbf{A complex framework may require more adaptation costs to fit difficult tasks.} As shown in Fig.~\ref{fig:code-agent-result}, although VerilogCoder continues to improve on \verilogevaltwo, its performance on \xname is not stable and may require more adjustments. Specifically, syntax correctness decreases as iterations progress. Analysis reveals that this may be because VerilogCoder gradually generates various subcomponents, so that in the beginning, the generated code is simpler and easier to ensure syntax correctness, and as the iterations progress, the code becomes increasingly complex, making it more difficult to guarantee syntax correctness.

\subsection{Failure Case Analysis}
In this section, we further analyze the factors that lead to failure, including submodules, input-output length, and circuit types.

\textbf{Submodule instantiation has a significant impact on the pass rate.} We select LLMs with higher performance and compare their pass rate in the task with and without submodules. 
The results in Table~\ref{tab:submodule} show that the formal@1 of tasks without submodules is 13.0\% higher than that of tasks with submodules on average, with DS-V3-671B showing the largest difference of 20.9\%.
It is also shown that LLMs' performance on tasks with submodules is similarly low, indicating that the ability to handle submodule relationships is a possible future research direction for achieving real-world designs.

\textbf{LLMs struggle with FSMs and complex combinational circuits.} We categorize the circuits in \xname into three types: combinational circuits, sequential circuits without finite state machines (FSMs), and sequential circuits with FSMs. 
The results in Table~\ref{tab:submodule} show that LLMs perform best in sequential circuits without FSMs (Seq w/o FSM), followed by combinational circuits (Comb), and worst in sequential circuits with FSMs (Seq w FSM). 
Further analysis reveals that the underperformance of combinational circuits is due to the fact that combinational circuits in the CPU mainly involve strict detail requirements, such as CSR control and long pipeline write-back, which are difficult for LLMs to process.

\begin{figure}[t]
\centerline{\includegraphics[width=\linewidth]{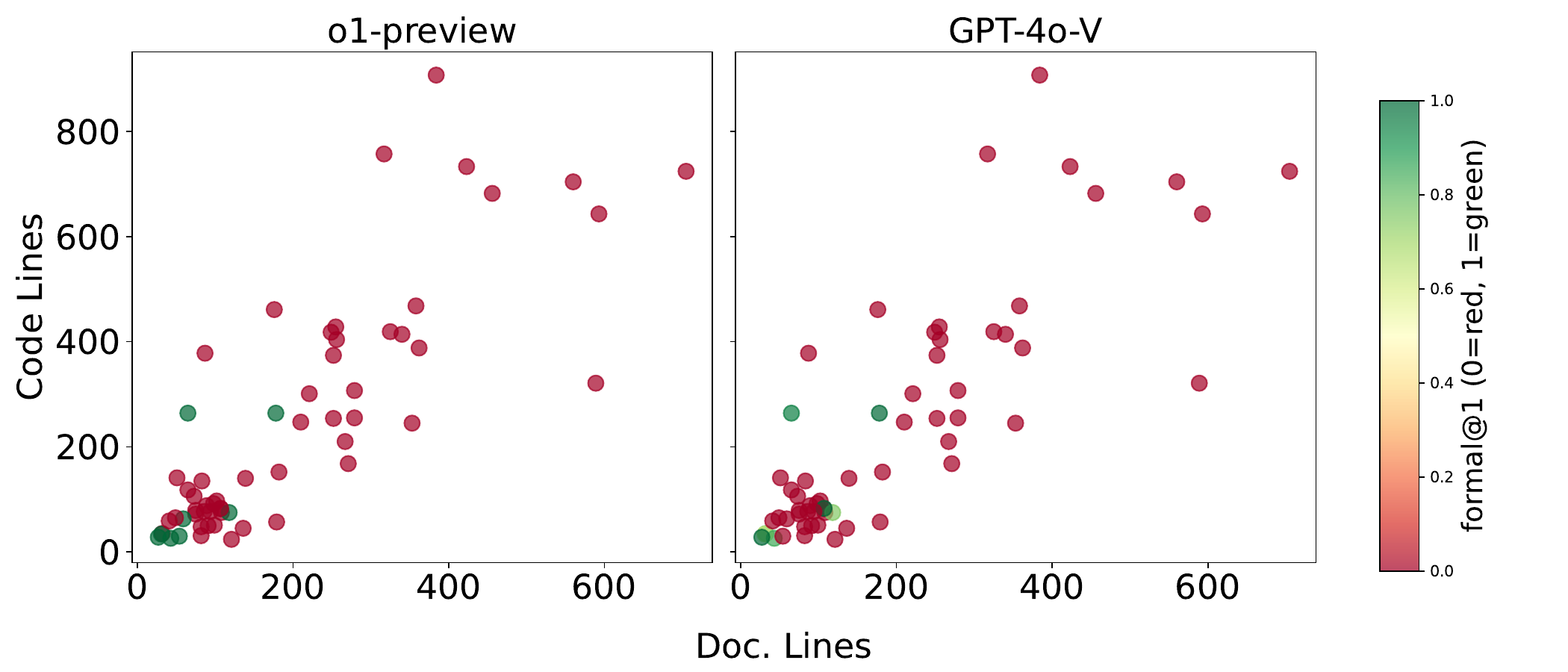}}
\caption{The pass rate of o1-preview and GPT-4o-V varies with the number of document lines and code lines. Each point represents a task, with red indicating a pass rate close to 0 and green indicating a pass rate close to 1.}
\label{fig:scatter}
\vspace{-10pt}
\end{figure}

\textbf{LLMs perform better with shorter input and output.} We analyze the results of o1-preview and GPT-4o-V. As shown in Fig.~\ref{fig:scatter}, LLMs can generate the correct modules for tasks with short design specifications and code, except for \texttt{aes\_sbox} and \texttt{aes\_inv\_sbox}, which have long code and design specifications but consist of simple data mapping between input and output signals.

%% file: sections/4_relatedwork.tex
\section{Related Work}

\subsection{Current Verilog Generation Benchmarks}

VeriGen~\cite{thakur2023benchmarking} pioneered the evaluation of LLM-generated Verilog code, though limited by its small-scale circuit coverage and insufficient testing methodology. 
\verilogeval~\cite{liu2023verilogeval} advanced the field with 156 designs and the golden model for functionality verification, while \rtllmone~\cite{lu2024rtllm} introduced comprehensive evaluation metrics including syntax, functionality, and design performance. 
Further developments include \verilogevaltwo~\cite{pinckney2024revisiting}, which incorporated in-context learning capabilities and systematic failure analysis.
\rtllmtwo~\cite{liu2024openllm} expands \rtllmone to 50 RTL designs across various design types, though still focusing on basic functional blocks rather than complex hierarchical systems.
Chang et al.~\cite{chang2024natural} introduced a novel multi-modal framework and related benchmarks, combining natural language with visual block diagrams.
RTL-Repo~\cite{allam2024rtl} addressed this limitation by providing over 4,000 samples for multi-file, large-context scenario assessment. 
SpecLLM~\cite{li2024specllm} proposes a benchmark for generation and review specifications with LLMs.

Compared to these benchmarks, \xname has significant advantages in terms of design complexity and reality, standardized design specifications, and rigorous verification, which can greatly contribute to the development of Verilog generation and hardware automation design community.

\subsection{LLMs for Verilog Generation}

Various approaches have been proposed to utilize LLMs for RTL generation, which is constrained by high-quality datasets~\cite{thakur2023benchmarking, liu2023rtlcoder, liu2023chipnemo, touvron2023llama, pei2024betterv, gao2024autovcoder, cui2024origen, zhang2024mg, zhao2024codev, liu2024craftrtl}. 
Typically, RTLCoder~\cite{liu2023rtlcoder} leveraged GPT to synthesize instruction-code pairs to create a fine-tuning dataset and fine-tune their LLMs.
BetterV~\cite{pei2024betterv} augmented its GitHub-sourced Verilog dataset through Verilog to C translation.
MG-Verilog~\cite{zhang2024mg} and CodeV~\cite{zhao2024codev} introduced multi-grained description-code datasets that vary in complexity and detail.
CraftRTL~\cite{liu2024craftrtl} identified challenges in synthetic data generation for non-text representations and performance variability during benchmark training, and introduced a targeted fine-tuning dataset for LLMs, addressing issues in Karnaugh maps, state transition diagrams, and waveform problems in Verilog code.

\subsection{LLM Frameworks for Verilog Generation}

Some research focuses on developing generation frameworks for Verilog tasks to enhance the performance of general-purpose LLMs~\cite{Blocklove_2023, chang2023chipgpt, lu2024rtllm, pinckney2024revisiting, fu2023gpt4aigchip, yan2024assertllm, tsai2024rtlfixer, sami2024aivril, verilogcoder}. 
Among them, Chip-Chat~\cite{Blocklove_2023} and ChipGPT~\cite{chang2023chipgpt} employ multi-round EDA and human feedback to aid LLMs in Verilog generation, but the need for extensive human intervention limits automation. 
RTLFixer~\cite{tsai2024rtlfixer} integrates retrieval-augmented generation (RAG) and ReAct to correct syntax errors in Verilog code. 
VerilogCoder~\cite{verilogcoder} developed a multi-agent system that can write Verilog and utilize tools to correct syntax and functional errors. 

We evaluate typical LLMs and frameworks for Verilog generation on \xname, and the results show that current methods are still unable to effectively accomplish complex and real-world Verilog generation tasks.

%% file: sections/5_conclusion.tex
\section{Conclusion}
In this paper, we propose \xname, a Verilog generation benchmark designed for real-world design workflows. \xname includes (1) complex designs from open-source IPs, (2) handwritten detailed and formatted design specifications, and (3) rigorous verification environments. 
Evaluations of various models on \xname indicate that current models still have limitations in Verilog generation and highlight several potential research directions: formal verification of LLM-generated code, less hallucinated reasoning models for complex tasks, and stronger models that can handle diagram inputs and submodule instantiations effectively.

%% file: conference_101719.bbl
\begin{thebibliography}{10}
\providecommand{\url}[1]{#1}
\csname url@samestyle\endcsname
\providecommand{\newblock}{\relax}
\providecommand{\bibinfo}[2]{#2}
\providecommand{\BIBentrySTDinterwordspacing}{\spaceskip=0pt\relax}
\providecommand{\BIBentryALTinterwordstretchfactor}{4}
\providecommand{\BIBentryALTinterwordspacing}{\spaceskip=\fontdimen2\font plus
\BIBentryALTinterwordstretchfactor\fontdimen3\font minus \fontdimen4\font\relax}
\providecommand{\BIBforeignlanguage}[2]{{%
\expandafter\ifx\csname l@#1\endcsname\relax
\typeout{** WARNING: IEEEtran.bst: No hyphenation pattern has been}%
\typeout{** loaded for the language `#1'. Using the pattern for}%
\typeout{** the default language instead.}%
\else
\language=\csname l@#1\endcsname
\fi
#2}}
\providecommand{\BIBdecl}{\relax}
\BIBdecl

\bibitem{patterson2012better}
D.~Patterson, ``For better or worse, benchmarks shape a field,'' \emph{Communications of the ACM}, vol.~55, 2012.

\bibitem{achiam2023gpt}
J.~Achiam, S.~Adler, S.~Agarwal, L.~Ahmad, I.~Akkaya, F.~L. Aleman, D.~Almeida, J.~Altenschmidt, S.~Altman, S.~Anadkat \emph{et~al.}, ``Gpt-4 technical report,'' \emph{arXiv preprint arXiv:2303.08774}, 2023.

\bibitem{liu2023chipnemo}
M.~Liu, T.-D. Ene, R.~Kirby, C.~Cheng, N.~Pinckney, R.~Liang, J.~Alben, H.~Anand, S.~Banerjee, I.~Bayraktaroglu \emph{et~al.}, ``Chipnemo: Domain-adapted llms for chip design,'' \emph{arXiv preprint arXiv:2311.00176}, 2023.

\bibitem{liu2023rtlcoder}
S.~Liu, W.~Fang, Y.~Lu, Q.~Zhang, H.~Zhang \emph{et~al.}, ``Rtlcoder: Outperforming gpt-3.5 in design rtl generation with our open-source dataset and lightweight solution,'' \emph{arXiv preprint arXiv:2312.08617}, 2023.

\bibitem{pei2024betterv}
Z.~Pei, H.-L. Zhen, M.~Yuan, Y.~Huang, and B.~Yu, ``Betterv: Controlled verilog generation with discriminative guidance,'' \emph{arXiv preprint arXiv:2402.03375}, 2024.

\bibitem{zhang2024mg}
Y.~Zhang, Z.~Yu, Y.~Fu, C.~Wan \emph{et~al.}, ``Mg-verilog: Multi-grained dataset towards enhanced llm-assisted verilog generation,'' \emph{arXiv preprint arXiv:2407.01910}, 2024.

\bibitem{zhao2024codev}
Y.~Zhao, D.~Huang, C.~Li, P.~Jin, Z.~Nan \emph{et~al.}, ``Codev: Empowering llms for verilog generation through multi-level summarization,'' \emph{arXiv preprint arXiv:2407.10424}, 2024.

\bibitem{dataisallyouneed}
K.~Chang, K.~Wang, N.~Yang, Y.~Wang, D.~Jin, W.~Zhu, Z.~Chen, C.~Li, H.~Yan, Y.~Zhou \emph{et~al.}, ``Data is all you need: Finetuning llms for chip design via an automated design-data augmentation framework,'' \emph{arXiv preprint arXiv:2403.11202}, 2024.

\bibitem{gao2024autovcoder}
M.~Gao, J.~Zhao, Z.~Lin, W.~Ding, X.~Hou, Y.~Feng, C.~Li, and M.~Guo, ``Autovcoder: A systematic framework for automated verilog code generation using llms,'' \emph{arXiv preprint arXiv:2407.18333}, 2024.

\bibitem{liu2024craftrtl}
M.~Liu, Y.-D. Tsai, W.~Zhou, and H.~Ren, ``Craftrtl: High-quality synthetic data generation for verilog code models with correct-by-construction non-textual representations and targeted code repair,'' \emph{arXiv preprint arXiv:2409.12993}, 2024.

\bibitem{sami2024aivril}
H.~Sami, P.-E. Gaillardon, V.~Tenace \emph{et~al.}, ``Aivril: Ai-driven rtl generation with verification in-the-loop,'' \emph{arXiv preprint arXiv:2409.11411}, 2024.

\bibitem{chang2023chipgpt}
K.~Chang, Y.~Wang, H.~Ren, M.~Wang, S.~Liang \emph{et~al.}, ``Chipgpt: How far are we from natural language hardware design,'' 2023.

\bibitem{Blocklove_2023}
J.~Blocklove, S.~Garg, R.~Karri, and H.~Pearce, ``Chip-chat: Challenges and opportunities in conversational hardware design,'' in \emph{2023 {ACM}/{IEEE} 5th Workshop on Machine Learning for {CAD} ({MLCAD})}.\hskip 1em plus 0.5em minus 0.4em\relax {IEEE}, sep 2023.

\bibitem{yao2024rtlrewriter}
X.~Yao, Y.~Wang, X.~Li, Y.~Lian, R.~Chen, L.~Chen, M.~Yuan, H.~Xu, and B.~Yu, ``Rtlrewriter: Methodologies for large models aided rtl code optimization,'' \emph{arXiv preprint arXiv:2409.11414}, 2024.

\bibitem{chen2024dawn}
L.~Chen, Y.~Chen, Z.~Chu, W.~Fang, T.-Y. Ho \emph{et~al.}, ``The dawn of ai-native eda: Promises and challenges of large circuit models,'' \emph{arXiv preprint arXiv:2403.07257}, 2024.

\bibitem{yan2024assertllm}
Z.~Yan, W.~Fang, M.~Li, M.~Li, S.~Liu, Z.~Xie, and H.~Zhang, ``Assertllm: Generating and evaluating hardware verification assertions from design specifications via multi-llms,'' \emph{arXiv preprint arXiv:2402.00386}, 2024.

\bibitem{fu2023gpt4aigchip}
Y.~Fu, Y.~Zhang, Z.~Yu, S.~Li, Z.~Ye, C.~Li, C.~Wan, and Y.~C. Lin, ``Gpt4aigchip: Towards next-generation ai accelerator design automation via large language models,'' in \emph{2023 IEEE/ACM International Conference on Computer Aided Design (ICCAD)}.\hskip 1em plus 0.5em minus 0.4em\relax IEEE, 2023, pp. 1--9.

\bibitem{wu2024chateda}
H.~Wu, Z.~He, X.~Zhang, X.~Yao, S.~Zheng \emph{et~al.}, ``Chateda: A large language model powered autonomous agent for eda,'' \emph{IEEE Transactions on Computer-Aided Design of Integrated Circuits and Systems}, 2024.

\bibitem{kande2023llm}
R.~Kande, H.~Pearce, B.~Tan, B.~Dolan-Gavitt, S.~Thakur, R.~Karri, and J.~Rajendran, ``Llm-assisted generation of hardware assertions,'' \emph{arXiv preprint arXiv:2306.14027}, 2023.

\bibitem{tsai2024rtlfixer}
Y.~Tsai, M.~Liu, and H.~Ren, ``Rtlfixer: Automatically fixing rtl syntax errors with large language model,'' in \emph{Proceedings of the 61st ACM/IEEE Design Automation Conference}, 2024, pp. 1--6.

\bibitem{yao2024hdldebugger}
X.~Yao, H.~Li, T.~H. Chan, W.~Xiao, M.~Yuan \emph{et~al.}, ``Hdldebugger: Streamlining hdl debugging with large language models,'' \emph{arXiv preprint arXiv:2403.11671}, 2024.

\bibitem{yan2023viability}
Z.~Yan, Y.~Qin, X.~S. Hu, and Y.~Shi, ``On the viability of using llms for sw/hw co-design: An example in designing cim dnn accelerators,'' in \emph{2023 IEEE 36th International System-on-Chip Conference (SOCC)}.\hskip 1em plus 0.5em minus 0.4em\relax IEEE, 2023, pp. 1--6.

\bibitem{liu2024openllm}
S.~Liu, Y.~Lu, W.~Fang, M.~Li, and Z.~Xie, ``Openllm-rtl: Open dataset and benchmark for llm-aided design rtl generation,'' 2024.

\bibitem{thakur2023benchmarking}
S.~Thakur, B.~Ahmad, Z.~Fan, H.~Pearce, B.~Tan, R.~Karri, B.~Dolan-Gavitt, and S.~Garg, ``Benchmarking large language models for automated verilog rtl code generation,'' in \emph{2023 Design, Automation \& Test in Europe Conference \& Exhibition (DATE)}.\hskip 1em plus 0.5em minus 0.4em\relax IEEE, 2023, pp. 1--6.

\bibitem{liu2023verilogeval}
M.~Liu, N.~Pinckney, B.~Khailany, and H.~Ren, ``Verilogeval: Evaluating large language models for verilog code generation,'' in \emph{2023 IEEE/ACM International Conference on Computer Aided Design (ICCAD)}.\hskip 1em plus 0.5em minus 0.4em\relax IEEE, 2023, pp. 1--8.

\bibitem{lu2024rtllm}
Y.~Lu, S.~Liu, Q.~Zhang, and Z.~Xie, ``Rtllm: An open-source benchmark for design rtl generation with large language model,'' in \emph{2024 29th Asia and South Pacific Design Automation Conference (ASP-DAC)}.\hskip 1em plus 0.5em minus 0.4em\relax IEEE, 2024, pp. 722--727.

\bibitem{chang2024natural}
K.~Chang, Z.~Chen, Y.~Zhou, W.~Zhu, H.~Xu \emph{et~al.}, ``Natural language is not enough: Benchmarking multi-modal generative ai for verilog generation,'' \emph{arXiv preprint arXiv:2407.08473}, 2024.

\bibitem{pinckney2024revisiting}
N.~Pinckney, C.~Batten, M.~Liu, H.~Ren, and B.~Khailany, ``Revisiting verilogeval: Newer llms, in-context learning, and specification-to-rtl tasks,'' \emph{arXiv preprint arXiv:2408.11053}, 2024.

\bibitem{allam2024rtl}
A.~Allam and M.~Shalan, ``Rtl-repo: A benchmark for evaluating llms on large-scale rtl design projects,'' \emph{arXiv preprint arXiv:2405.17378}, 2024.

\bibitem{li2024specllm}
M.~Li, W.~Fang, Q.~Zhang, and Z.~Xie, ``Specllm: Exploring generation and review of vlsi design specification with large language model,'' \emph{arXiv preprint arXiv:2401.13266}, 2024.

\bibitem{opencores}
OpenCores, ``Opencores: Open source ip-cores,'' https://opencores.org/.

\bibitem{opencores-ds}
------, ``Design sdc\_mmc controller,'' https://github.com/freecores/sdcard \_mass\_storage\_controller.

\bibitem{e203_hbirdv2}
riscv mcu, ``Hummingbirdv2 e203 core and soc,'' https://github.com/riscv-mcu/e203\_hbirdv2/tree/master.

\bibitem{jimenez2024swebench}
\BIBentryALTinterwordspacing
C.~E. Jimenez, J.~Yang, A.~Wettig, S.~Yao, K.~Pei, O.~Press, and K.~R. Narasimhan, ``{SWE}-bench: Can language models resolve real-world github issues?'' in \emph{The Twelfth International Conference on Learning Representations}, 2024. [Online]. Available: \url{https://openreview.net/forum?id=VTF8yNQM66}
\BIBentrySTDinterwordspacing

\bibitem{yang2024swe}
J.~Yang, C.~Jimenez, A.~Wettig, K.~Lieret, S.~Yao, K.~Narasimhan, and O.~Press, ``Swe-agent: Agent-computer interfaces enable automated software engineering,'' \emph{Advances in Neural Information Processing Systems}, vol.~37, pp. 50\,528--50\,652, 2024.

\bibitem{Yosys}
C.~Wolf, ``Yosys open synthesis suite,'' \url{https://yosyshq.net/yosys/}.

\bibitem{Nangate}
Silvaco, ``Nangate freepdk45 generic open cell library,'' https://si2.org/open-cell-library/.

\bibitem{jaspergold}
Cadence, ``Jasper formal verification platform,'' https://www.cadence. com/en\_US/home/tools/system-design-and-verification/formal-and-static-verification/jasper-verification-platform.html.

\bibitem{lin2004rouge}
C.-Y. Lin, ``Rouge: A package for automatic evaluation of summaries,'' in \emph{Text summarization branches out}, 2004, pp. 74--81.

\bibitem{dubey2024llama}
A.~Dubey, A.~Jauhri, A.~Pandey, A.~Kadian, A.~Al-Dahle \emph{et~al.}, ``The llama 3 herd of models,'' \emph{arXiv preprint arXiv:2407.21783}, 2024.

\bibitem{deepseekv3}
A.~Liu, B.~Feng, B.~Xue, B.~Wang, B.~Wu \emph{et~al.}, ``Deepseek-v3 technical report,'' \emph{arXiv preprint arXiv:2412.19437}, 2024.

\bibitem{deepseekr1}
D.~Guo, D.~Yang, H.~Zhang, J.~Song, R.~Zhang \emph{et~al.}, ``Deepseek-r1: Incentivizing reasoning capability in llms via reinforcement learning,'' \emph{arXiv preprint arXiv:2501.12948}, 2025.

\bibitem{deepseekcoder}
D.~Guo, Q.~Zhu, D.~Yang, Z.~Xie, K.~Dong \emph{et~al.}, ``Deepseek-coder: When the large language model meets programming--the rise of code intelligence,'' \emph{arXiv preprint arXiv:2401.14196}, 2024.

\bibitem{codeqwen}
\BIBentryALTinterwordspacing
Q.~Team, ``Code with codeqwen1.5,'' April 2024. [Online]. Available: \url{https://qwenlm.github.io/blog/codeqwen1.5/}
\BIBentrySTDinterwordspacing

\bibitem{verilogcoder}
C.-T. Ho, H.~Ren, and B.~Khailany, ``Verilogcoder: Autonomous verilog coding agents with graph-based planning and abstract syntax tree (ast)-based waveform tracing tool,'' in \emph{Proceedings of the AAAI Conference on Artificial Intelligence}, vol.~39, no.~1, 2025, pp. 300--307.

\bibitem{touvron2023llama}
H.~Touvron, L.~Martin, K.~Stone, P.~Albert, A.~Almahairi \emph{et~al.}, ``Llama 2: Open foundation and fine-tuned chat models,'' \emph{arXiv preprint arXiv:2307.09288}, 2023.

\bibitem{cui2024origen}
F.~Cui, C.~Yin, K.~Zhou, Y.~Xiao, G.~Sun \emph{et~al.}, ``Origen: Enhancing rtl code generation with code-to-code augmentation and self-reflection,'' in \emph{Proceedings of the 43rd IEEE/ACM International Conference on Computer-Aided Design}, 2024, pp. 1--9.

\end{thebibliography}
